\documentclass[11pt,letterpaper]{article}

\pdfoutput=1

\usepackage[T1]{fontenc}
\usepackage[utf8]{inputenc}
\usepackage{times}
\usepackage{latexsym}
\usepackage{inconsolata}

\usepackage{microtype}

\usepackage{amsmath}
\usepackage{amssymb}
\usepackage{amsthm}

\usepackage{algorithm}
\usepackage{algpseudocode}

\usepackage{booktabs}
\usepackage{multirow}
\usepackage{graphicx} 
\usepackage{graphicx}
\usepackage{tikz}
\usepackage{tabularx} 

\usepackage{xcolor}

\usepackage{hyperref}

\usepackage{natbib}

\usepackage{xcolor}
\usepackage{array} 
\usepackage{tabularx} 
\usepackage{multirow}  
\usepackage{float}

\usepackage[final]{acl2023}

\title{PatentMind: A Multi-Aspect Reasoning Graph\\for Patent Similarity Evaluation}

\author{
Yongmin Yoo, \quad Qiongkai Xu, \quad Longbing Cao \\
Frontier AI Research Centre, Macquarie University \\
School of Computing, FSE, Macquarie University\\
\texttt{yooyongmin91@gmail.com} \quad 
\texttt{\{qiongkai.xu, longbing.cao\}@mq.edu.au}
}

\begin{document}
\maketitle
\begin{abstract}
Patent similarity evaluation plays a critical role in intellectual property analysis. However, existing methods often overlook the intricate structure of patent documents, which integrate technical specifications, legal boundaries, and application contexts. We introduce PatentMind, a novel framework for patent similarity assessment based on a Multi-Aspect Reasoning Graph (MARG). PatentMind decomposes patents into their three dimensions of technical features, application domains, and claim scopes, then dimension-specific similarity scores are calculated over the MARG. These scores are dynamically weighted through a context-aware reasoning process, which integrates contextual signals to emulate expert-level judgment. To support evaluation, we construct a human-annotated benchmark PatentSimBench, comprising 500 patent pairs. Experimental results demonstrate that the PatentMind-generated scores show a strong correlation ($r=0.938$) with expert annotations, significantly outperforming embedding-based models, patent-specific models, and advanced prompt engineering methods. Beyond computational linguistics, our framework provides a structured and semantically grounded foundation for real-world decision-making, particularly for tasks such as infringement risk assessment, underscoring its broader impact on both patent analytics and evaluation.
\end{abstract}

\section{Introduction}

Patent documents pose significant challenges for NLP-based similarity evaluation due to specialized domain knowledge, intricate legal language, and complex structural formats~\citep{aslanyan2022patents,indukuri2007similarity,casola2021summarization,lupu2013patent}. Accurate evaluation of patent similarity is crucial for identifying prior art, assessing infringement risks, and determining the novelty of innovations~\citep{Feng2020,Helmers2019,arts2018text,Wang2023}. Given that inaccuracies can lead to overlooked prior art, invalid patent grants, or undetected infringements, robust and interpretable evaluation methods are essential. Moreover, the rapidly growing volume of patent applications across diverse technological fields further intensifies the need for reliable similarity evaluation techniques~\citep{jiang2024nlp}. Given the increasing reliance on patent analytics for R\&D strategy, legal risk assessment, and competitive intelligence, developing accurate and interpretable similarity models has become critically important in both industry and academia~\citep{hain2022similarity,yoo2023novel}.
 
\begin{figure}[t]
  \centering
  \includegraphics[width=0.99\linewidth,keepaspectratio]{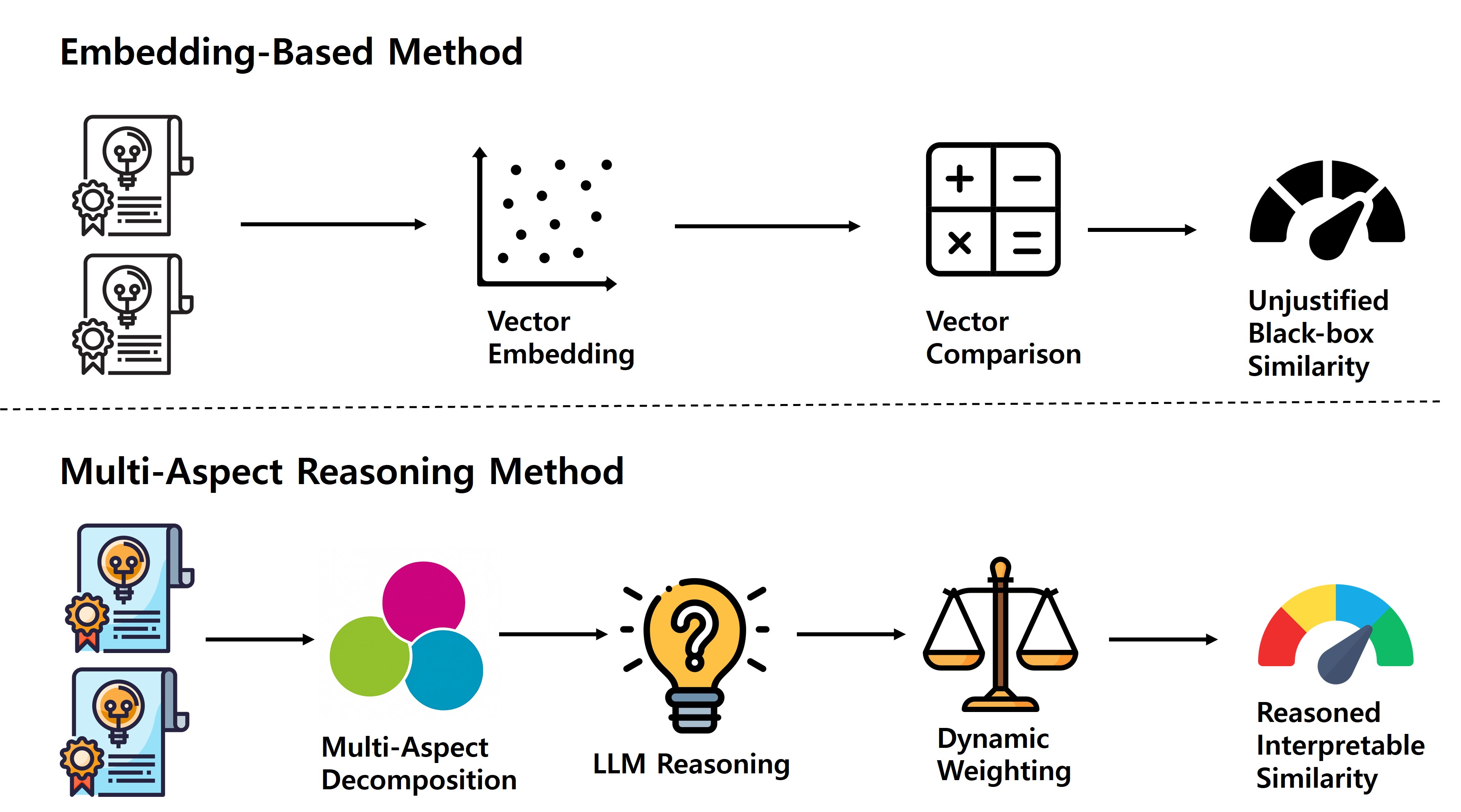}
  \caption{The comparison between multi-aspect reasoning and embedding-based similarity methods.}
  \label{fig:intro_overview}
  \vspace{-4mm}
\end{figure}

In the early stages of patent similarity evaluation, keyword-based methods such as bag-of-words were widely used. However, these approaches struggled to capture the semantics and technical specificity of patent documents, resulting in ineffective representations~\citep{D'hondt2013,ascione2024comparative}. To overcome these limitations, embedding-based methods such as static word embeddings and transformer-based contextual embeddings were subsequently introduced~\citep{lee2020patentbert,reimers2019sentence}. Nevertheless, these methods compress patent texts into dense vectors, which can obscure critical distinctions and reduce interpretability.

Recent advances in large language models (LLMs) have demonstrated considerable promise in reasoning-based tasks, offering a compelling alternative to traditional embedding-based approaches. Unlike fixed embedding methods, LLMs support dynamic reasoning and semantic comprehension, facilitating more sophisticated similarity assessments. However, directly applying general-purpose LLMs to patent similarity tasks remains challenging due to specialized terminology and complex features inherent in patent texts~\citep{ikoma2025can,ascione2024comparative}. Moreover, prompt-based approaches to LLMs have partially addressed the limitations of embeddings, but still struggle due to the specialized terminology and structural complexity of patent documents.

To address these challenges, we propose PatentMind, a novel patent similarity evaluation framework structured as a Multi-Aspect Reasoning Graph (MARG). This framework emulates the analytical processes of patent experts through multi-step reasoning based on LLMs. PatentMind decomposes patent documents into three core dimensions: technical features, application domains, and claim scopes. Each dimension is independently evaluated for similarity, and the results are integrated through a context-aware reasoning process consisting of four stages: Domain Relationship Analysis, Information Distribution, Dimension Relevance, and Cross-Validation. This multi-dimensional approach is designed to closely reflect real-world patent examination practices.~\citep{uspto_mpep_2141}.

Consequently, PatentMind effectively captures subtle distinctions across the diverse attributes of patents. Our empirical evaluations demonstrate that PatentMind, implemented with GPT-4o-mini as the underlying LLM, achieves a Pearson correlation of 0.938 with expert evaluations, significantly surpassing embedding and LLM-based prompting baselines. To facilitate rigorous evaluation, we introduce PatentSimBench as the first expert-annotated patent similarity benchmark that reflects real-world criteria employed by patent examiners and attorneys. By explicitly modeling technical, legal, and contextual aspects through interpretable reasoning, PatentMind collectively address key limitations of prior approaches.

The key contributions of PatentMind are summarized as follows:

\begin{itemize}
    \item \textbf{Multi-Aspect Reasoning Graph (MARG)}: We propose a structured LLM framework that performs modular similarity evaluation across three core dimensions of patent documents and integrates them through interpretable, multi-step reasoning.
    \item \textbf{Context-Aware Reasoning}: We introduce a dynamic weighting mechanism built on multi-step contextual analysis, incorporating domain relationship, score distribution, dimension relevance, and robustness validation.
    \item \textbf{Performance Superiority}: We demonstrate that PatentMind consistently surpasses embedding-based, prompting-based, and even fine-tuned models that require substantial time and computational resources, with this advantage arising from its patent-specialized, multi-dimensional reasoning framework rather than mere reliance on LLM calls.
    \item \textbf{PatentSimBench Dataset}: We release the first publicly available expert-annotated benchmark for patent similarity evaluation, comprising 500 high-quality patent pairs. 
\end{itemize}
\section{Related Work}
\subsection{Patent Similarity Evaluation}

\label{sec:methodology}
\begin{figure*}[t]
    \centering    
    \includegraphics[width=0.99\textwidth,keepaspectratio]{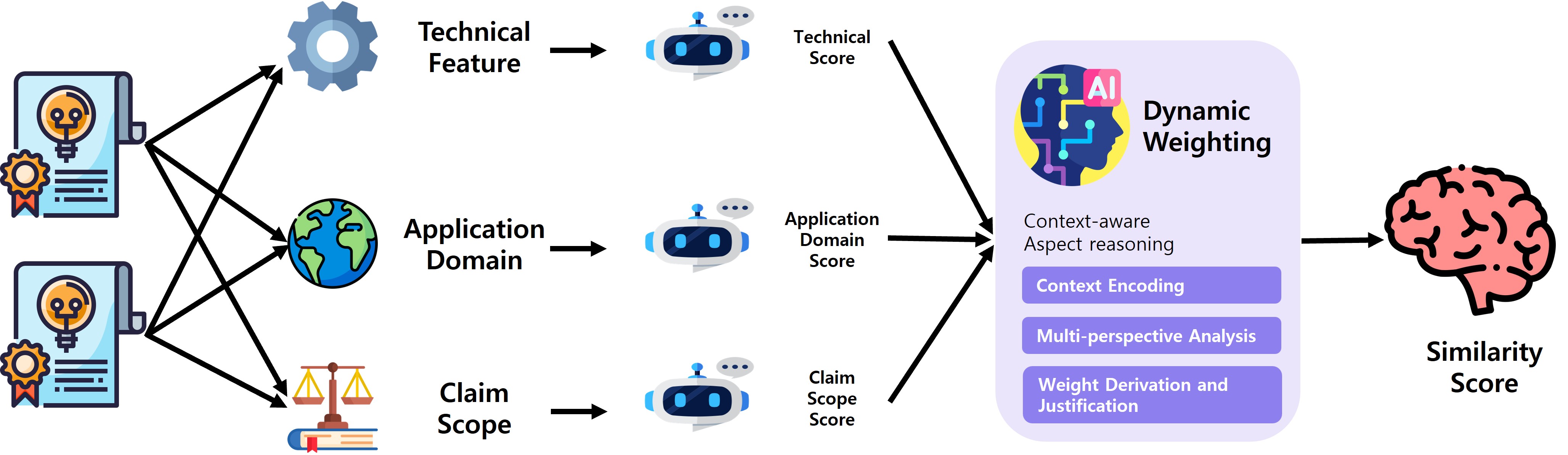}
    \caption{The workflow of PatentMind, structured as a Multi-Aspect Reasoning Graph (MARG).}
    \label{fig:patentmind_workflow}
    \vspace{-4mm}
\end{figure*}

Early approaches to patent similarity evaluation, such as keyword matching and TF-IDF, treated patents as generic texts and often failed to capture their domain-specific structure and terminology~\citep{tseng2007text,ascione2024comparative}. To address these limitations, vector-based models like Word2Vec and Doc2Vec were introduced to encode semantic relationships within patent corpora~\citep{jeon2022doc2vec}. However, these models struggled with challenges such as polysemy and limited domain adaptation.
To improve semantic representation, contextualized language models like SciBERT~\citep{beltagy2019scibert} and PatentBERT~\citep{lee2020patentbert} were developed, offering embeddings tailored for scientific and patent texts. While these models produce more accurate representations, their black-box nature and lack of interpretability hinder their practical application in legal and technical decision-making contexts~\citep{miric2023using,ascione2024comparative}.

To overcome these challenges, we propose PatentMind, a structured and interpretable framework that mimics expert reasoning via modular, dimension-based analysis. By decomposing patents into multiple perspectives and reasoning over each dimension, PatentMind provides transparent, multi-dimensional similarity scores aligned with human judgment criteria, addressing the limitations of opaque, single-vector embeddings.

\subsection{Prompt Engineering}
Prompt engineering has emerged as a key strategy for enabling complex reasoning in LLMs. Chain-of-Thought prompting~\citep{wei2022chain} improves sequential reasoning by guiding models through intermediate steps. However, its linear structure limits effectiveness in tasks requiring divergent reasoning. To overcome this, Yao et al.~\citep{yao2023tree} proposed the Tree-of-Thoughts framework, which enables exploration of multiple reasoning paths in parallel, enhancing performance on tasks requiring structured deliberation. Additional work demonstrates that carefully designed prompts enable strong few-shot~\citep{brown2020language} and zero-shot~\citep{kojima2022large} reasoning, using minimal task examples or simple cues. These prompting strategies significantly improve LLM performance on complex tasks, laying the foundation for PatentMind's structured reasoning approach, where multi-dimensional analysis is essential for capturing both technical and legal dimensions of patents.




\section{Methodology}

This section presents PatentMind, a novel framework for patent similarity evaluation structured as a Multi-Aspect Reasoning Graph (MARG). PatentMind systematically leverages the reasoning capabilities of LLMs to analyze patent documents across multiple critical dimensions. The framework decomposes patent texts into distinct dimensions, independently evaluates each dimension, and integrates the resulting similarity scores through a context-aware dynamic weighting mechanism. Figure~\ref{fig:patentmind_workflow} illustrates the MARG workflow, consisting of three core modules: Multi-Aspect Decomposition, Similarity Computation, and Context-Aware Dynamic Weighting.

\subsection{Multi-Dimensional Feature Extraction}
\label{subsec:feature_extraction}
PatentMind captures the complex structure and domain-specific nuances of patent documents through LLM-based feature extraction. We decompose each patent into three dimensions: technical feature (\(T\)), application domain (\(D\)), and claim scope (\(C\)). These dimensions reflect how patent specialists evaluate patent similarity, focusing on what the invention is, where it applies, and how broadly it is claimed \citep{uspto_mpep_2141}.

To obtain these dimensions, we apply an LLM-based feature extraction function, \(T\), \(D\), or \(C\), on a patent document \(P\) to obtain corresponding content. The function generates a structured representation composed of technical features \(T(P)\), application domains \(D(P)\), and claim scope \(C(P)\).
The title, abstract, and claims of the patent are used as input for the extractor. Note that these sections are legally mandated components for any valid patent application (e.g., 35 U.S.C. § 112), ensuring the availability of structured inputs. This structured prompting approach guides the LLM using aspect-specific instructions. Each prompt focuses on a distinct semantic aspect: methodologies for technical features, application contexts for domains, and legal boundaries for claim scope. These prompts are grounded in structured inputs from the title, abstract, and claims. Prompt designs for all dimensions are provided in Appendix~\ref{app:prompts_feature_extraction}.

\begin{figure*}[t]
    \centering    
    \includegraphics[width=0.99\textwidth,keepaspectratio]{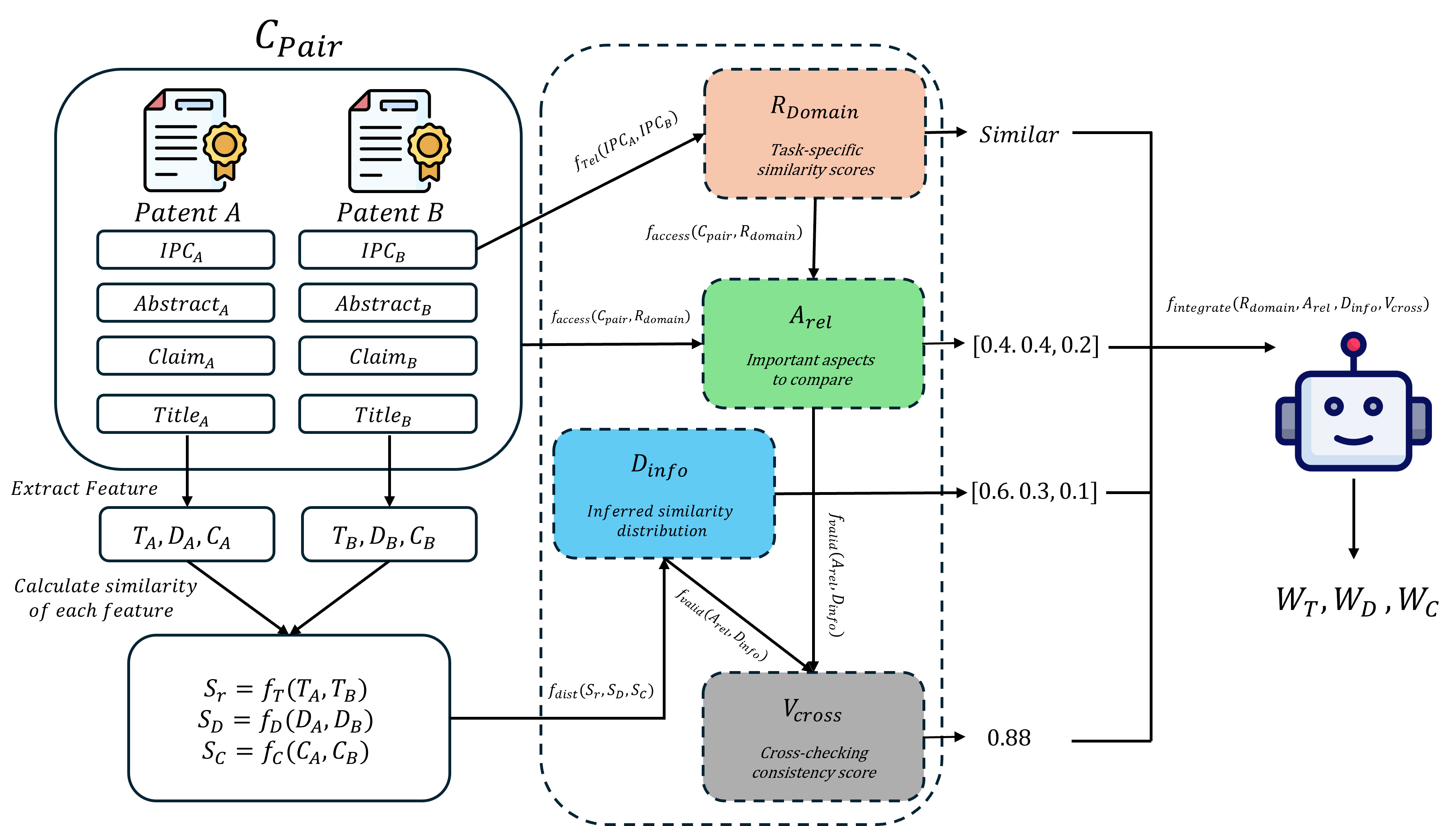}
    \caption{The computational graph of the Context-Aware Dynamic Weight Reasoning framework, with modules annotated by their notations: Domain Relationship Analysis ($f_{rel}, R_{domain}$), Information Distribution Analysis ($f_{dist}, D_{info}$), Dimension Relevance Assessment ($f_{assess}, A_{rel}$), and Cross-validation Reasoning ($f_{valid}, V_{cross}$).}
    \label{fig:MARG_workflow}
    \vspace{-4mm}
\end{figure*}

\subsection{Dimension-wise Similarity Computation}
\label{subsec:similarity_computation}

Following feature extraction, we compute similarity between patent documents on these aspects. Given two patent documents $ P_1 $ and $ P_2 $ and their respective extracted features $ T(P_i) $, $ D(P_i) $, and $ C(P_i) $, we calculate similarity scores across each dimension:
\begin{equation}
S = \{ S_T, S_D, S_C \},
\end{equation}
where $S_T$, $S_D$, and $S_C$ denote the similarity scores for the technical, domain, and claim dimensions, respectively, computed as $S_T = f_T(T(P_1), T(P_2))$, $S_D = f_D(D(P_1), D(P_2))$, and $S_C = f_C(C(P_1), C(P_2))$. These scores serve as the foundation for the final weighted similarity measure described in Section~\ref{subsec:dynamic_weighting}.

We implement similarity functions through targeted prompts that direct the LLM to assess similarity using semantic relationships, contextual implications, and domain-specific equivalences. Each similarity score ranges from 0 to 1, with higher values indicating greater similarity. Appendix \ref{app:prompts_similarity_computation} details the prompt templates used.

\subsection{Context-Aware Dynamic Weight Reasoning}
\label{subsec:dynamic_weighting}

PatentMind adapts to contextual nuances by computing weights through multi-step LLM reasoning. Instead of parameter-intensive fine-tuning, it functions as an open calculation process using reasoning prompts, highlighting both efficiency and adaptability. This process comprises four key stages.

\subsubsection{Context Encoding}
We construct a contextual representation of the patent pair by encoding key information from their titles, abstracts, claims, and IPC codes:
\begin{equation}
C_{pair} = \{ t, a, c, IPC \}_{A,B},
\end{equation}
where $ t $, $ a $, $ c $, and $ IPC $ represent the title, abstract, claims, and $ IPC $ codes of patents $ A $ and $ B $.

\subsubsection{Multi-perspective Analysis}
The LLM performs multi-dimensional reasoning to guide the dynamic weighting process through four interdependent functions (detailed prompts in Appendix~\ref{app:prompts}).
First, \textbf{\textit{Domain Relationship Analysis}} ($f_{rel}$) leverages IPC codes to output a categorical label $R_{domain}$, determining the influence of domain similarity.
Second, \textbf{\textit{Information Distribution Analysis}} ($f_{dist}$) examines the patterns of aspect-wise scores ($S_T, S_D, S_C$) to generate $D_{info}$.
Third, \textbf{\textit{Dimension Relevance Assessment}} ($f_{assess}$) evaluates dimension importance ($A_{rel}$) based on the full context $C_{pair}$ and $R_{domain}$.
Finally, \textbf{\textit{Cross-validation Reasoning}} ($f_{valid}$) checks the consistency between predicted relevance ($A_{rel}$) and actual score distribution ($D_{info}$), yielding a robustness score $V_{cross} \in [0,1]$.
We formalize this pipeline as a directed acyclic graph $G=(V, E)$, where nodes correspond to reasoning steps and edges represent dependencies facilitating the message-passing of contextual signals for the final weight derivation.

\subsubsection{Weight Derivation and Justification}
In this step, the integration function $f_{integrate}$ computes the final dimension weights $w_T$, $w_D$, and $w_C$ by aggregating the outputs of the reasoning modules: the domain relationship label ($R_{domain}$), the information distribution signal ($D_{info}$), the dimension relevance scores ($A_{rel}$), and the robustness indicator ($V_{cross}$). It also generates a textual justification $J$ that explains the rationale behind the derived weights, thereby enhancing interpretability. Detailed prompting instructions for each reasoning step, including $f_{integrate}$, are provided in Appendix~\ref{app:prompts}.

\subsection{Final Similarity Calculation}
The overall similarity score combines the scores by
\begin{equation}
S_{\text{overall}} = w_T \times S_T + w_D \times S_D + w_C \times S_C .
\end{equation}
Here, the weights are normalized such that $w_T + w_D + w_C = 1$.
This weighted integration enables PatentMind to adaptively emphasize the most informative dimensions of each patent pair, reflecting the nuanced evaluation process of human patent experts. Appendix~\ref{app:similarity_calculation} details the prompts for this final step. 
\section{Dataset Construction}
\label{sec:dataset_construction}

To facilitate rigorous evaluation of patent similarity methods, we introduce PatentSimBench, the first expert-annotated benchmark dataset tailored for patent similarity tasks. PatentSimBench provides rationale-supported similarity scores aligned with real-world legal and technical judgments, making it particularly suitable for evaluating multi-step reasoning models. The dataset consists of 500 patent pairs sampled from the USPTO database~\citep{uspto_bulk_data}, with a balanced representation across all IPC sections. The International Patent Classification (IPC) categorizes patents into eight top-level domains, such as human necessities, chemistry, physics, and digital technologies~\citep{wipo_ipc}, ensuring comprehensive technical coverage across diverse fields. Similarity annotations are based on a 5-point Likert scale, with scores ranging from 1 to 5. Our dataset is publicly available on GitHub(It will be available upon accept).

\subsection{Dataset Composition}
\label{subsec:dataset_composition}

\begin{table}[t]
    \centering
    \footnotesize 
    \setlength{\tabcolsep}{8pt} 
    \begin{tabular}{lc}
        \toprule
        \textbf{Attribute} & \textbf{Value} \\
        \midrule
        Patent Pairs & 500 pairs \\
        Technical Fields & IPC Sections \\
        Similarity Range & 1 to 5 (Likert scale) \\
        \midrule
        Fleiss' Kappa & 0.588 \\
        Cronbach's Alpha & 0.967 \\   
        \bottomrule
    \end{tabular}
    \caption{The statistics of PatentSimBench.}
    \label{tab:dataset_stats}
\end{table}

PatentSimBench includes patent pairs evenly distributed across major technological domains, covering all IPC sections. Table~\ref{tab:dataset_stats} summarizes key dataset statistics. The moderate inter-annotator agreement scores, including Fleiss' Kappa (0.588) and Cronbach's Alpha (0.967), demonstrate robust consensus among expert annotators despite the inherent subjectivity in assessing patent similarity.

\subsection{Annotation Process}
Four domain experts with both legal and technical backgrounds annotated PatentSimBench using a 5-point Likert scale. The annotation guidelines were rigorously designed based on the USPTO Manual of Patent Examining Procedure (MPEP) \S 2141, which mandates a structured analysis of claim scope, technical content, and field of search. This ensures that our similarity criteria reflect actual legal standards rather than subjective judgments. The annotation process followed a two-phase protocol: (1) independent assessments by each expert, and (2) consensus refinement through structured discussions of divergent cases, during which annotators reviewed and adjusted their ratings based on shared rationales. For quality control, we excluded patent pairs with a post-discussion standard deviation greater than 2 (\(\sigma_{\text{ratings}} > 2\)), ensuring high reliability in the final dataset. This process was guided by the predefined annotation guidelines described in Appendix~\ref{app:annotation_guidelines}.

\section{Experimental Results}
\label{sec:experiments}

We conducted comprehensive experiments on the PatentSimBench dataset to evaluate PatentMind against embedding-based models, prompt engineering strategies, and regression baselines. To assess the robustness and contribution of individual components, we performed ablation studies and cross-model evaluations using various LLMs. All experiments were carried out with standard evaluation metrics, using GPT-4o-mini as the backbone model. 
To ensure determinism and reproducibility, we used a fixed temperature of 0.2 and default sampling parameters (top\_p=1.0).  Furthermore, to verify result stability, we conducted 5 independent runs with different random seeds on a representative subset of the data. 
The observed standard deviation in Pearson correlation was negligible ($\sigma < 0.001$), confirming the robustness of our results without the need for reporting averaged metrics over multiple trials.

\subsection{Comparison with Embedding-Based and Patent specific Methods}
\label{subsec:baseline_comparison}
\begin{table}[h]
\centering
\tiny 
\setlength{\tabcolsep}{1pt} 
\renewcommand{\arraystretch}{.8} 
\resizebox{.45\textwidth}{!}{ 
\begin{tabular}{lcccc}
\toprule
\textbf{Model} & \textbf{Pearson ($r$)} $\uparrow$ & \textbf{Spearman ($\rho$)} $\uparrow$ & \textbf{MSE} $\downarrow$ & \textbf{MAE} $\downarrow$ \\
\midrule
Word2Vec & .761 & .819 & .145 & .319 \\
BERT & .835 & .819 & .228 & .381 \\
BERT-for-Patent & .762 & .754 & .115 & .283 \\
SciBERT & .766 & .798 & .194 & .381 \\
Patent-GPT-J & .892 & .883 & .124 & .178 \\
\textbf{PatentMind} & \textbf{.938} & \textbf{.923} & \textbf{.113} & \textbf{.092} \\
\bottomrule
\end{tabular}
}
\caption{The performance comparison of PatentMind with varying embedding-based and patent specific baseline methods.}
\label{tab:baseline_comparison}
\end{table}

Table~\ref{tab:baseline_comparison} compares PatentMind with conventional embedding-based and patent-specific models. Baseline approaches exhibit only moderate correlation with expert judgments and show high error rates. In contrast, PatentMind achieves significantly better performance with a Pearson correlation of 0.938 and the lowest error rates, demonstrating superior alignment with human evaluations.

\subsection{Comparison with Prompting-Based Methods}
\label{subsec:prompting_comparison}

\begin{table}[h]
\centering
\scalebox{1.15}{ 
\scriptsize
\setlength{\tabcolsep}{3.9pt}
\begin{tabular}{lcc}
\toprule
\textbf{Prompting Strategy} & \textbf{Pearson ($r$)} $\uparrow$ & \textbf{Spearman ($\rho$)} $\uparrow$ \\
\midrule
I/O Prompting & .702 & .672 \\
Few-shot Prompting & .891 & .897 \\
Chain-of-Thought (CoT) & .866 & .870 \\
Self-Consistency w/ CoT & .887 & .881 \\
Tree-of-Thought (ToT) & .745 & .654 \\
Chain-of-Draft (CoD) & .834 & .835 \\
\textbf{PatentMind} & \textbf{.938} & \textbf{.923} \\
\bottomrule
\end{tabular}
}
\caption{The performance comparison of PatentMind with various prompting-based baseline methods.}
\label{tab:prompting_comparison}
\end{table}

We compare PatentMind with representative prompting strategies in Table~\ref{tab:prompting_comparison}. While advanced methods such as Few-shot, Chain-of-Thought (CoT), and Self-Consistency yield reasonable performance, they struggle to capture the multi-dimensional nature of patent documents. For instance, CoT typically follows a single reasoning path, limiting its ability to jointly model technical, legal, and domain-specific aspects. Self-Consistency introduces variation in reasoning but fails to incorporate context-sensitive judgment aligned with expert reasoning. In contrast, PatentMind integrates hierarchical reasoning with dynamic weighting, enabling more faithful modeling of patent semantics. As a result, it achieves the highest agreement with expert judgments (Pearson $r = .938$; Spearman $\rho = .923$). These findings highlight the critical role of explicitly modeling multi-aspect patent contexts in enhancing both accuracy and interpretability.

\subsection{Comparison of LLMs}
\label{subsec:llm_evaluation}

\begin{table}[h]
\centering
\tiny
\setlength{\tabcolsep}{1.8pt} 
\renewcommand{\arraystretch}{.81}
\resizebox{.45\textwidth}{!}{
\begin{tabular}{lcc}
\toprule
\textbf{LLM} & \textbf{Pearson ($r$)} $\uparrow$ & \textbf{Spearman ($\rho$)} $\uparrow$ \\
\midrule
Claude-3.5-Sonnet & .931 & .928 \\
Llama-3.3-70B-Versatile-128k & .922 & .911 \\
Deepseek-r1-distill-llama-70b & .914 & .907 \\
Qwen-QwQ-32B-Preview& .910 & .893 \\
GPT-4o-mini(PatentMind) & \textbf{.938} & \textbf{.923} \\
\bottomrule
\end{tabular}
}
\caption{The performance comparison of PatentMind using various LLMs.}
\label{tab:llm_evaluation}
\end{table}

To test PatentMind’s robustness across LLM architectures, we evaluated its performance with multiple models. All tested LLMs showed high correlation with expert annotations, indicating that PatentMind’s effectiveness is not model-dependent. GPT-4o-mini and Claude-3.5-Sonnet performed competitively, while Qwen-QwQ-32B showed slightly lower scores, likely due to architectural and reasoning differences. Overall, the consistent results highlight the model-agnostic strength of PatentMind’s design.

\subsection{Comparison of Regression Weighting}

\begin{table}[h]
\centering
\scalebox{.65}{
\begin{tabular}{lcccc}
\toprule
\textbf{Fold Type} & \textbf{Pearson} & \textbf{Spearman} & \textbf{RMSE} & \textbf{MAE} \\
\midrule
Linear Regression  & .920 & .897 & .122 & .095 \\
Lasso Regression & .918 & .901 & .127 & .104 \\
SVM & .919 & .895 & .125 & .102 \\
Tree Boosting & .909 & .886 & .129 & .097 \\
MLP Regressor & .921 & .896 & .121 & .095 \\
Bayesian Ridge Regression & .920 & .897 & .122 & .096 \\
\textbf{PatentMind} & \textbf{.938} & \textbf{.923} & \textbf{.113} & \textbf{.092} \\
\bottomrule
\end{tabular}
}
\caption{The performance comparison of PatentMind using dynamic weighting and regression based baselines.}
\label{tab:comparison}
\end{table}

To validate the effectiveness of PatentMind's dynamic weighting strategy, we compared it against several regression-based baselines using the PatentSimBench dataset. Model performance was evaluated based on Pearson and Spearman correlations between predicted similarity scores and expert annotations. As shown in Table~\ref{tab:comparison}, all regression models, including Linear Regression, Lasso, SVM, Tree Boosting, Bayesian Ridge, and a 2-layer MLP Regressor, underperformed relative to PatentMind. The best-performing baseline, MLP Regression (Pearson $r = .921$; Spearman $\rho = .896$), still fell short of PatentMind’s performance (Pearson $r = .938$; Spearman $\rho = .923$). These results reveal that regression-based fixed-weighting methods struggle to capture the nuanced, context-sensitive variations essential for accurate patent similarity evaluation. Unlike these baselines, which require supervised training to learn fixed parameters, PatentMind functions as an open reasoning process driven by LLM prompts, avoiding extra training costs while dynamically adapting weights to each patent pair.

\subsection{Efficiency and Cost-Benefit Analysis}
To assess the practical viability of PatentMind, we evaluated its computational efficiency using GPT-4o-mini. 
On average, processing a single patent pair requires approximately 13,225 input tokens and 832 output tokens across 8 API calls. 
Through parallelized execution of independent extraction and scoring steps, the end-to-end latency is reduced to 20--30 seconds per pair. Financially, the operational cost is approximately \$0.003 USD per pair based on current API pricing. 
While computationally more expensive than simple embedding-based retrieval, this cost is negligible compared to human patent experts. 
As shown in Table~\ref{tab:efficiency}, PatentMind offers expert-level reasoning at a fraction of the cost ($<$\$0.01), making it a cost-effective solution for high-stakes decision-making tasks such as infringement risk assessment.

\begin{table}[h]
\centering
\small
\caption{Efficiency and cost comparison. Human cost is estimated based on average patent attorney hourly rates assuming 60 minutes per case.}
\label{tab:efficiency}
\begin{tabular}{lcc}
\toprule
Method & Latency & Cost / Pair \\
\midrule
Embedding (SciBERT) & $<$ 0.1s & $\approx$ \$0.000 \\
PatentMind (Ours) & 20~30s & $\approx$ \$0.003 \\
Human Expert & $>$ 60 min & $>$ \$100.0 \\
\bottomrule
\end{tabular}
\end{table}

We emphasize that this computational overhead is an intentional design choice. 
PatentMind is explicitly designed to function as a re-ranking and precision analysis module, typically applied after a broad initial retrieval phase. 
In high-stakes domains like patent infringement analysis or invalidity searches, where missing a critical prior art can result in significant legal consequences, accuracy and interpretability take precedence over millisecond latency. 
Therefore, the trade-off of higher computational cost for expert-level reasoning depth is justified for these targeted downstream applications.

\section{Ablation Studies}
\subsection{Comparison of Aspect Impact}
\label{subsec:ablation}

\begin{table}[ht]
\centering
\scalebox{.9}{  
\scriptsize
\begin{tabular}{lccc}
\toprule
\textbf{Model} & \textbf{Pearson} & \textbf{Spearman} & \textbf{Avg. Drop (\%)} \\
\midrule
\textbf{PatentMind Full} & \textbf{.938} & \textbf{.923} & -- \\
Equal Weighting & .904  & .890  & \textbf{3.35} \\
\midrule
\multicolumn{4}{l}{\textbf{Two Dimensions Only}} \\
\quad w/o Claim Scope Dimension & .912  & .906  & 2.15 \\
\quad w/o Technical Dimension & .901  & .898  & 3.10 \\
\quad w/o Application Dimension & .903  & .894  & 3.20 \\
\textbf{Average (Two Dimension)} & \textbf{.905} & \textbf{.899} & \textbf{2.82} \\
\midrule
\multicolumn{4}{l}{\textbf{Single Dimension Only}} \\
\quad Claim Scope Dimension Only & .903 & .905  & 2.65 \\
\quad Technical Dimension Only & .908  & .912  & 2.05 \\
\quad Application Dimension Only & .865  & .848  & 7.40 \\
\textbf{Average (Single Dimension)} & \textbf{.892} & \textbf{.888} & \textbf{4.03} \\
\bottomrule
\end{tabular}
}
\caption{The performance comparison of PatentMind using varying aspects.}
\label{tab:ablation_avg_drop}
\end{table}

To assess the contribution of each component in our framework, we conducted an ablation study across seven configurations, including the removal of dynamic weighting, exclusion of individual dimensions, and single-dimension-only variants. As shown in Table~\ref{tab:ablation_avg_drop}, replacing the context-aware dynamic weighting mechanism with equal weighting results in a noticeable performance drop (3.35\%), demonstrating the importance of adapting weights to contextual relevance. Removing any one of the three core dimensions(Claim Scope, Technical, or Application)also led to performance degradation (avg. 2.82\%), indicating their complementary roles in similarity evaluation. Notably, the removal of Technical (3.10\%) or Application (3.20\%) dimensions caused greater performance loss than removing Claim Scope (2.15\%). Single-dimension variants showed even larger drops in performance (avg. 4.03\%), with the Application-only model suffering the most (7.40\%). Although the Technical-only model performed relatively better (2.05\% drop), it still fell short of the full model's performance. These findings validate the effectiveness of our design choices: the integration of technical, legal, and domain-level reasoning, combined with context-aware dynamic weighting, leads to substantially improved accuracy and adaptability in patent similarity evaluation.

\section{Error Analysis}
\label{sec:error}

In this section, we examine the performance boundaries of PatentMind. We analyze prediction accuracy and residual patterns, identify domain-specific error trends across IPC sections, and investigate high-error cases involving functional ambiguity and terminology mismatch. 

\subsection{Prediction and Residual Analysis}

\begin{figure}[ht]
    \centering
    \includegraphics[width=0.99\linewidth, keepaspectratio]{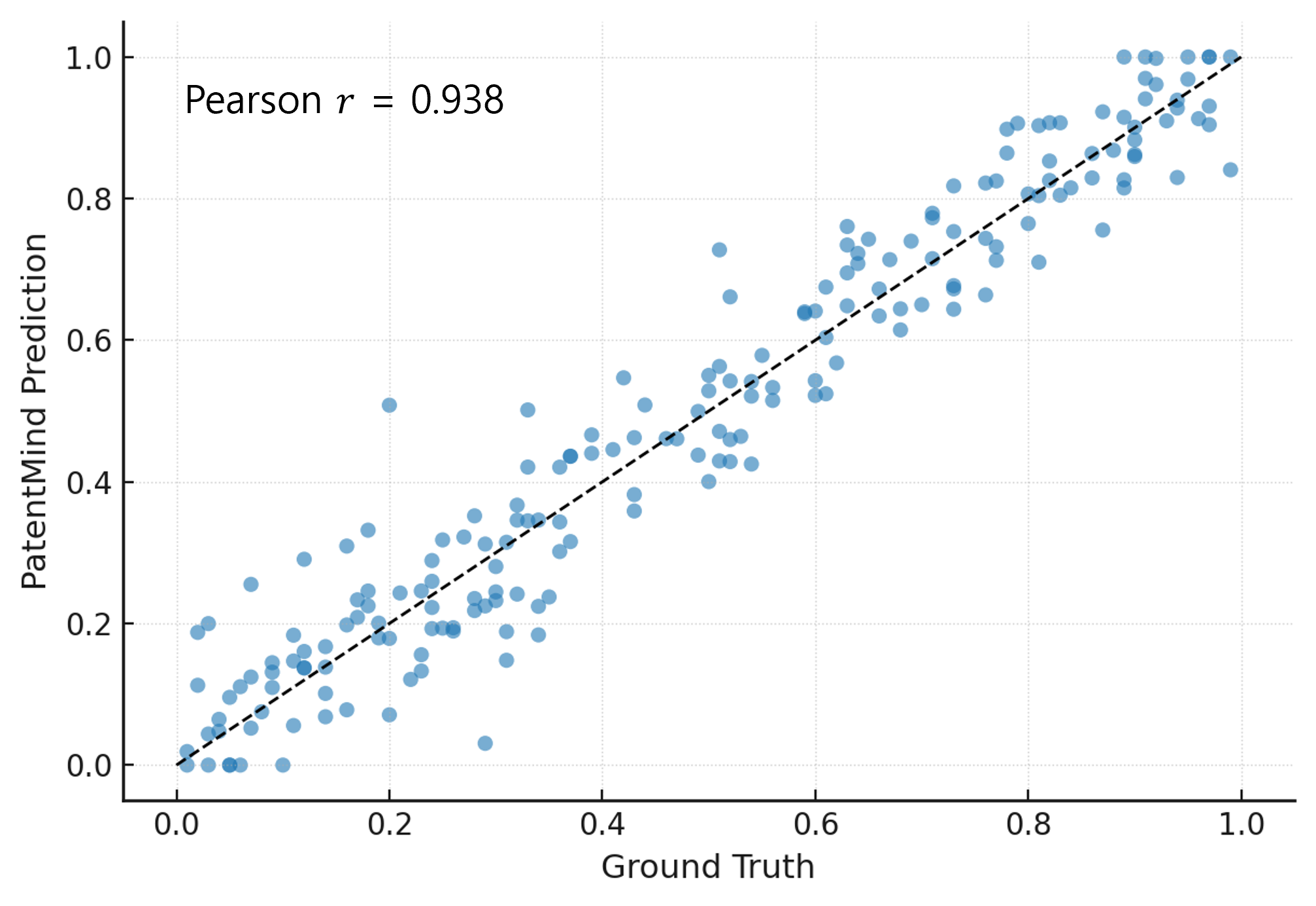}
    \caption{The correlation between expert judgments and scores predicted by PatentMind.}
    \label{fig:scatter}
    \vspace{-4mm}
\end{figure}

We assess PatentMind’s reliability by comparing predicted similarity scores to expert annotations and analyzing residuals. Predictions show strong alignment, with most errors within the 0--0.1 range (MAE = 0.0918). Figure~\ref{fig:scatter} shows predictions generally follow the ideal ($y = x$) line, though high-similarity pairs (ground truth $>0.8$) tend to be slightly underestimated, and low-similarity pairs ($<0.2$) slightly overestimated. The average residual ($-0.0197$) indicates minimal bias, with domain-specific variations discussed further below.

\subsection{Error Analysis across Different IPC Codes}
\label{subsec:error}
Our IPC-level analysis shows that PatentMind’s prediction errors are more pronounced in certain technical domains, such as metallurgy (C22B), genetic engineering (C12N), and mechanical engineering (F16D). These errors stem from: (1) underrepresentation of domain-specific terms in LLM pretraining (e.g., “leaching” vs. “extraction” in C22B), (2) structural variations in patent drafting across domains, and (3) context-dependent similarity criteria. For instance, metallurgy patents (e.g., C22B300) often suffer from underestimated similarity due to synonym recognition issues (e.g., “leaching” vs. “extraction”), while audio processing patents (e.g., G06F316) tend to be overestimated due to language. Biotechnology patents (C12N) present additional challenges due to linguistic complexity and causal reasoning structures.

These findings underscore the need for domain-adaptive embeddings or knowledge graph integration to enhance synonym recognition and contextual understanding. PatentMind’s modular design facilitates such enhancements without requiring extensive retraining, a key direction for future work.

\subsection{High-Error Case Analysis}

We further analyze the 50 patent pairs with the highest prediction errors to identify potential systematic limitations. We observe four recurring patterns:
(1) underestimation of highly similar patents (e.g., predicted 0.53 vs. actual 0.833 in metallurgy);
(2) overestimation of dissimilar patents (e.g., predicted 0.50 vs. actual 0.250 in marking tech);
(3) difficulty distinguishing functional differences despite textual overlap (e.g., 0.48 vs. 0.750 in mechanical parts);
(4) challenges in interpreting domain terms, particularly in biotech (e.g., 0.24 vs. 0.500).
These findings suggest that, while PatentMind effectively captures general semantic similarity, performance may be further improved by incorporating more explicit representations of functional intent and domain-specific vocabulary. Additional examples and discussion are provided in Appendix~\ref{appendix:high_error_cases}.

\section{Conclusion}
\label{sec:conclusion}

We proposed PatentMind, a novel framework integrating linguistic, legal, and technical reasoning for patent similarity. By jointly modeling technical features, application domains, and claim scope, it effectively captures structural complexity. A key innovation is our context-aware weighting mechanism, which uses LLM reasoning to dynamically adjust dimensional importance.

PatentMind achieves expert-level agreement ($r=0.938$) and model-agnostic robustness, significantly outperforming embedding, prompting, and patent-tuned baselines. Notably, it delivers higher accuracy with greater efficiency than resource-intensive fine-tuning approaches.

Finally, we release PatentSimBench, the first expert-annotated similarity benchmark. These contributions extend beyond computational linguistics, offering a semantically grounded foundation for critical real-world tasks such as prior art search, infringement assessment, and novelty evaluation. 

\section*{Limitations}
The PatentMind framework exhibits robust performance in patent similarity evaluation, surpassing prior methodologies with a Pearson correlation of 0.938 against expert annotations, yet presents specific limitations that merit future refinement without compromising its foundational contributions.

\paragraph{Computational Efficiency.}
PatentMind’s multi-stage reasoning structure and dynamic weighting mechanism deliver high precision and interpretability but introduce relatively high computational complexity, which may necessitate additional optimization for large-scale or latency-sensitive applications. This issue is distinct from PatentMind’s core research objective (emphasizing accuracy in patent similarity evaluation) and will be addressed in future work. Specifically, subsequent efforts will explore optimizations such as response caching, selective execution of reasoning steps, and lightweight model substitution (e.g., knowledge distillation or LoRA-based LLMs). These approaches are informed by recent studies demonstrating that LLM distillation techniques, such as the Distilling Step-by-Step method, can effectively transfer multi-stage reasoning capabilities to smaller models, and that LoRA’s low-rank adaptation is well-suited for dynamically optimizing model weights. Furthermore, PatentMind’s modular design enables flexible adaptation to efficiency requirements, and techniques like prompt compression and batch parallel processing (e.g., tensor parallelization or in-flight batching for optimized GPU utilization) can further enhance scalability in deployment environments.

These limitations do not diminish PatentMind’s primary contributions, including achieving a 0.938 Pearson correlation with expert annotations. Rather, they underscore the research’s sustainability and potential for advancement. By explicitly acknowledging these constraints and proposing concrete directions for improvement, this work establishes a long-term impact in the field of patent similarity evaluation.

\paragraph{Domain-Specific Opportunities.} While PatentMind demonstrates strong overall performance, our error analysis highlights opportunities for further enhancement across different technical domains. The current framework adopts a general approach, which provides broad applicability but can be further strengthened to capture domain-specific nuances in highly specialized fields. As shown in Section ~\ref{subsec:error}, domains with complex terminology (e.g., metallurgy, biotechnology) or unique structural conventions represent promising areas where targeted improvements may yield additional gains.

Our dynamic weighting mechanism already helps address these challenges by considering IPC code relationships, and future work can build on this by integrating explicit domain knowledge. Although the current evaluation spans diverse IPC sections, expanding dataset coverage will enable more fine-grained statistical insights at specific domain levels. This creates an opportunity to design more specialized adaptation strategies.

In particular, tailored prompts for high-error domains, domain-specific terminology processing, and fine-tuning approaches for technical fields can further amplify PatentMind’s effectiveness. Importantly, the modular design of PatentMind ensures that such domain-adaptive components can be integrated seamlessly without reengineering the overall architecture, underscoring its flexibility and extensibility for future advancements.

\section*{Ethical Considerations}
This research adheres to ethical principles in data utilization.  The PatentSimBench dataset comprises exclusively publicly available patent documents, complying with intellectual property regulations and containing no personally identifiable information.  Acknowledging potential biases in language models, we implemented expert annotations and structured reasoning methodologies to enhance the interpretability and fairness of the results.  To facilitate transparency and reproducibility, we have made our research methodology and dataset publicly accessible.  We have considered the potential misuse of AI-based patent similarity evaluation tools in legal disputes or competitive intelligence analysis.  Consequently, we recommend using this system as a supplementary tool rather than as the sole basis for high-stakes decisions.  Our future research will continue investigating bias mitigation strategies and enhancing ethical applications of patent evaluation systems. For transparency, we note that ChatGPT was used to improve grammar and clarity of expression.

\bibliography{anthology,custom}
\bibliographystyle{acl_natbib}

\appendix
\appendix

\appendix
\section{Appendix A: Patent Terminology and Concepts}
\label{appendix:patent_terminology}

This appendix provides detailed explanations of patent terminology and concepts referenced in the main paper, intended to help readers less familiar with intellectual property documentation.

\subsection{Patent Document Structure}

A patent document follows a standardized format consisting of several key components:

\begin{itemize}
    \item \textbf{Title:} A brief description of the invention that typically indicates its function, mechanism, or purpose.
    
    \item \textbf{Abstract:} A concise summary (typically 150-250 words) outlining the invention's core technical contribution.
    
    \item \textbf{Background:} Explanation of the technical problem being addressed, limitations of existing solutions, and the context of the invention.
    
    \item \textbf{Detailed Description:} Comprehensive explanation of the invention's implementation, often including drawings, diagrams, specific embodiments, and working examples sufficient to enable reproduction by a person skilled in the relevant field.
    
    \item \textbf{Claims:} Precisely defined statements that establish the legal boundaries of protection. Claims are the most critical section for legal analysis and similarity assessment.
\end{itemize}

Claims are structured hierarchically, with independent claims defining the broadest protection and dependent claims adding specific limitations. For example, an independent claim might describe a general method for data processing, while dependent claims specify particular implementations, parameters, or use cases.

\subsection{Patent Classification Systems}

The International Patent Classification (IPC) system organizes patents using hierarchical alphanumeric codes that categorize technological domains. The IPC structure includes:

\begin{itemize}
    \item \textbf{Section} (one letter, A-H): Broadest division of technology (e.g., A = Human Necessities, G = Physics)
    \item \textbf{Class} (two digits): Major technological divisions within a section (e.g., G06 = Computing)
    \item \textbf{Subclass} (one letter): Further division (e.g., G06F = Electric Digital Data Processing)
    \item \textbf{Group} (variable digits): Specific technological areas (e.g., G06F3/048 = Interaction techniques for graphical user interfaces)
\end{itemize}

Common IPC codes referenced in this paper include:
\begin{itemize}
    \item \textbf{C12N:} Biochemistry - Microorganisms or enzymes
    \item \textbf{G06F:} Computing - Electric digital data processing
    \item \textbf{F16D:} Mechanical engineering - Couplings and brakes
    \item \textbf{C22B:} Metallurgy - Production or refining of metals
\end{itemize}

Each patent may be assigned multiple IPC codes to reflect its cross-disciplinary nature. These classifications are crucial for organizing patent literature, identifying relevant prior art, and contextualizing similarity assessments.

\subsection{Legal Concepts in Patent Analysis}

Several legal terms are fundamental to patent analysis:

\begin{itemize}
    \item \textbf{Prior art:} Any evidence that an invention is already known before the filing date of a patent application. This includes existing patents, published applications, academic papers, public demonstrations, or commercial products. Prior art determines novelty and is a primary consideration in examining patent validity.
    
    \item \textbf{Novelty:} For an invention to be patentable, it must be new (novel) compared to prior art. A patent lacks novelty if all its essential elements are disclosed in a single prior art reference.
    
    \item \textbf{Non-obviousness:} Beyond novelty, an invention must involve an inventive step that would not be obvious to a person with ordinary skill in the relevant technical field.
    
    \item \textbf{Infringement:} Unauthorized making, using, selling, or importing of a patented invention. Infringement analysis examines whether a product or process incorporates all elements of at least one independent claim of a patent.
    
    \item \textbf{Claim construction:} The process of interpreting the meaning and scope of patent claims, which is essential for both infringement analysis and similarity assessment.
\end{itemize}

\subsection{Multi-dimensional Nature of Patent Similarity}

Patent similarity assessment is inherently multi-faceted, encompassing three critical dimensions that patent experts consider during evaluation:

\begin{itemize}
    \item \textbf{Technical attributes:} The core invention mechanisms, algorithms, components, or methodologies. This dimension focuses on how the invention works and what technical problems it solves. Technical similarity might exist even when patents are applied in different domains.
    
    \item \textbf{Application contexts:} The fields, industries, problems, or use cases where the invention applies. Two patents may implement different technical approaches but address the same application problem, resulting in contextual similarity.
    
    \item \textbf{Legal boundaries:} The scope and limitations of protection defined by the claims. Legal similarity assessment considers the overlap in protection scope, which may differ from pure technical similarity. Broader claims typically encompass more potential similar patents than narrowly defined claims.
\end{itemize}

Patent examiners, attorneys, and analysts dynamically adjust the importance of these dimensions based on the specific context of the analysis task. For example, prior art searches emphasize technical similarity to assess novelty, while infringement analysis focuses on claim coverage and legal boundaries.

This multi-dimensional nature makes patent similarity assessment particularly challenging and distinguishes it from general document similarity tasks, justifying our development of the MARG framework that explicitly models these dimensions.

\section{Appendix B: Prompts for Feature Extraction}
\label{app:prompts_feature_extraction}

In this appendix, we provide the prompts used to guide the Large Language Model (LLM) in extracting the three essential dimensions from patent documents: Technical Features (\( T(P) \)), Application Domains (\( D(P) \)), and Claim Scope (\( C(P) \)).

\subsection{Prompt for Technical Features (\( T(P) \))}
Prompt : \textit{Summarize the technical features of the patent, focusing on methodologies, algorithms, and innovation points.}

\subsection{Prompt for Application Domains (\( D(P) \))}
Prompt : \textit{Identify the application domains of the patent, including industries, problem areas, and potential applications.}

\subsection{Prompt for Claim Scope (\( C(P) \))}
Prompt : \textit{Determine the claim scope of the patent, summarizing the legal protection boundaries and key rights asserted in the claims.}

\section{Appendix C: Prompts for Similarity Computation}
\label{app:prompts_similarity_computation}

In this appendix, we provide the prompts used to guide the Large Language Model (LLM) in computing the similarity scores for each of the three dimensions: Technical Features (\( S_T \)), Application Domains (\( S_D \)), and Claim Scope (\( S_C \)). These prompts enable the LLM to assess the degree of overlap between two patent documents based on their extracted features.

\subsection{Prompt for Calculate Technical Features Similarity (\( S_T \))}
Prompt : \textit{Given the technical feature summaries of Patent A and Patent B, assess the similarity of their technical contributions, focusing on methodologies, algorithms, and innovation points. Provide a similarity score between 0 and 1, where 0 indicates no overlap and 1 indicates identical technical features. Include a brief justification for your assessment. Output the result in the following format: Score: [numerical score], Reason: [justification].}

\subsection{Prompt for Calculate Application Domains Similarity (\( S_D \))}
Prompt : \textit{Given the application domain summaries of Patent A and Patent B, evaluate the similarity of their practical contexts, including industries, problem areas, and potential applications. Provide a similarity score between 0 and 1, where 0 indicates completely distinct domains and 1 indicates fully shared domains. Include a brief justification for your assessment.Output the result in the following format: Score: [numerical score], Reason: [justification].}

\subsection{Prompt for Calculate Claim Scope Similarity (\( S_C \))}
Prompt : \textit{Given the claim scope summaries of Patent A and Patent B, analyze the similarity of their legal protection boundaries and key rights asserted in the claims. Provide a similarity score between 0 and 1, where 0 indicates no overlap in claim scope and 1 indicates identical claim scope. Include a brief justification for your assessment.Output the result in the following format: Score: [numerical score], Reason: [justification].}

\section{Appendix D: Prompt for Dynamic Weight}
\label{app:prompts}

Below are the prompts used to guide the Large Language Model (LLM) through the five sequential reasoning stages of the context-aware dynamic weighting process in the MAGR framework. Each prompt specifies the input data and requires a structured output format.

\subsection{Prompt for Domain Relationship Analysis (\( R_{domain} \))}
Prompt: \textit{Given the IPC codes of Patent A and Patent B from \( C_{pair} \) (titles, abstracts, claims, and IPC codes), assess the technical domain relationship between the two patents. Categorize the relationship as identical (same IPC subclass), hierarchical (one patent's domain subsumes the other), overlapping (shared IPC codes), or distinct (no common IPC codes). Output the result in the following format: Category: [relationship], Explanation: [justification].}

\subsection{Prompt for Information Distribution Analysis (\( D_{info} \))}
Prompt: \textit{Given the similarity scores \( S_T \), \( S_D \), and \( S_C \) for Patent A and Patent B, as computed in Section~\ref{subsec:similarity_computation}, analyze the distribution pattern of these scores. Identify the pattern as uniform similarity (all scores are similar), dimension dominance (one score is significantly higher), or complementary dimensions (high similarity in one dimension offsets lower similarity in others). Output the result in the following format: Pattern: [pattern], Justification: [explanation].}

\subsection{Prompt for Dimension Relevance Assessment (\( A_{rel} \))}
Prompt: \textit{Given the context of Patent A and Patent B from \( C_{pair} \) (titles, abstracts, claims, and IPC codes) and their domain relationship \( R_{domain} \), assess the relative importance of technical features, application domains, and claim scope for evaluating their similarity. Assign relevance scores between 0 and 1 to each dimension, ensuring the sum equals 1. Output the result in the following format: Scores: [technical features: score, application domains: score, claim scope: score], Explanation: [justification].}

\subsection{Prompt for Cross-validation Reasoning (\( V_{cross} \))}
Prompt: \textit{Given the dimension relevance scores (\(A_{rel}\)) and the actual similarity score distribution (\(D_{info}\)), assess how well these two align. If they strongly agree (i.e., the most important predicted dimension matches the dimension with the highest similarity), assign a robustness score close to 1. If they partially agree or conflict, assign a lower robustness score accordingly. Output the result in the following format: Metric: [score], Justification: [explanation].}

\subsection{Prompt for Weight Derivation and Justification (\protect\( w_T, w_D, w_C, J \protect\))}

Prompt: \textit{Given the domain relationship $R_{domain}$, similarity distribution pattern $D_{info}$, relevance scores $A_{rel}$, and robustness metric $V_{cross}$ for Patent A and Patent B, integrate these inputs to determine the final weights for technical features $w_T$, application domains $w_D$, and claim scope $w_C$, ensuring $w_T + w_D + w_C = 1$. Provide a textual justification. Output the result in the following format: Weights: [$w_T$: score, $w_D$: score, $w_C$: score], Justification: [explanation].}

\section{Appendix E: Prompt for Similarity Score Calculation}
\label{app:similarity_calculation}
    Prompt : \textit{Calculate the final similarity score \( S_{\text{final}} \) using the formula \( S_{\text{final}} = w_T \times S_T  + w_D \times S_D + w_C \times S_C \), where \( S_T \), \( S_D \), and \( S_C \) are the similarity scores for technical, application domains, and claim scope, respectively, and \( w_T \), \( w_D \), and \( w_C \) are their corresponding weights. Ensure that the result is a numerical value between 0 and 1, and return the value rounded to three decimal places. Output the result in the following format. Patent\_Similarity\_MAR :[score]}

\section{Appendix F: PatentSimBench Annotation Guidelines}
\label{app:annotation_guidelines}

In this appendix, we provide the annotation guidelines used for constructing the PatentSimBench dataset. These guidelines were distributed to all expert annotators to ensure consistent evaluation of patent similarity.

\subsection{Project Introduction}
PatentSimBench is the first expert-annotated benchmark dataset for evaluating similarity between patent documents. The purpose of this project is to establish a reliable ``gold standard'' for patent similarity across various technical domains.

\subsection{Rating Scale}
Annotators evaluated patent pair similarity using a 5-point Likert scale:

\begin{table}[ht]
\centering
\scriptsize
\setlength{\tabcolsep}{3pt}
\renewcommand{\arraystretch}{1.7}
\begin{tabular}{ccl}
\toprule
\textbf{Score} & \textbf{Level} & \textbf{Description} \\
\midrule
1 & Very Low & Fundamentally different inventions \\
2 & Low & Similar elements with substantial differences \\
3 & Medium & Partial overlap in key dimensions \\
4 & High & Substantial similarity in core technology \\
5 & Very High & Nearly identical inventions \\
\bottomrule
\end{tabular}
\caption{Similarity Rating Scale for Patent Pairs}
\label{tab:rating_scale}
\end{table}

\subsection{Similarity Assessment Guidelines}
Annotators were instructed to consider the following elements when evaluating similarity between patents:

\subsubsection{Core Invention Concept}
\textbf{Definition}: The essence of the problem being solved and its solution.

\textbf{Assessment Guidelines}:
\begin{itemize}
    \item Fundamental problem-solving approach in both patents
    \item Core concepts and principles of the inventions
    \item Key innovation points and technical contributions
    \item Fundamental similarities and differences between inventions
\end{itemize}

\subsubsection{Implementation Details}
\textbf{Definition}: Specific methods and components that realize the core concept.

\textbf{Assessment Guidelines}:
\begin{itemize}
    \item Physical/logical components and their arrangement
    \item Implementation details and mechanisms
    \item Performance parameters and operating conditions
    \item Comparison across various embodiments of the invention
\end{itemize}

\subsubsection{Purpose and Effects}
\textbf{Definition}: Intended effects and purposes of the invention.

\textbf{Assessment Guidelines}:
\begin{itemize}
    \item Benefits and effects provided by the invention
    \item Intended usage environment and situations
    \item Degree of problem resolution achieved by the invention
    \item Impact on users or industries
\end{itemize}

\subsubsection{Legal Protection Dimensions}
\textbf{Definition}: The scope of legal protection sought by the patent document.

\textbf{Assessment Guidelines}:
\begin{itemize}
    \item Scope and limitations of claims
    \item Characteristics of key rights assertions
    \item Clarity and specificity of legal protection
    \item Legal differentiation from existing patents
\end{itemize}

\subsection{Annotation Procedure}
Annotators followed a structured process:
\begin{enumerate}
    \item Preliminary review of both patent documents
    \item Structural analysis of each patent according to assessment guidelines
    \item Comparative analysis to identify similarities and differences
    \item Assessment of similarity across all evaluation elements
    \item Determination of the overall similarity score (1-5)
    \item Documentation of detailed rationale (minimum 100 words)
\end{enumerate}

\subsection{Assessment Principles}
Annotators were guided by the following principles:
\begin{itemize}
    \item \textbf{Objectivity}: Focus on document content rather than personal preferences
    \item \textbf{Consistency}: Assign consistent scores to patent pairs with similar characteristics
    \item \textbf{Detailed Review}: Analyze core technical/legal dimensions rather than superficial similarity
    \item \textbf{Patent Law Perspective}: Apply evaluation approaches based on USPTO MPEP guidelines
\end{itemize}

\subsection{Examples and References}

\subsubsection{Example 1: High Similarity (Score 5)}
\textbf{Patent Pair}: Patent A and Patent B (both describing a specific method for neural network acceleration)

\textbf{Key Similarities}:
\begin{itemize}
    \item Both patents address the same problem of neural network computation acceleration
    \item Identical core technical approach using matrix decomposition
    \item Similar architectural components and data flow
    \item Equivalent performance claims and applications in mobile devices
    \item Nearly identical claim scope with minor variations in dependent claims
\end{itemize}

\textbf{Annotator Rationale}: "These patents are nearly identical in their core innovation, implementation approach, and intended applications. Both describe the same matrix decomposition technique for neural network acceleration, with the same architectural components and data flow patterns. While Patent B includes two additional dependent claims specifying memory management details, this represents a minor extension rather than a fundamental difference. The technical overlap is comprehensive, and the claims protect essentially the same invention."

\subsubsection{Example 2: Medium Similarity (Score 3)}
\textbf{Patent Pair}: Patent C and Patent D (both related to image processing systems)

\textbf{Key Similarities/Differences}:
\begin{itemize}
    \item Both patents address image processing, but for different applications (medical imaging vs. autonomous vehicles)
    \item Similar preprocessing techniques but different core algorithms
    \item Partial overlap in component architecture but significant differences in implementation
    \item Different performance metrics and optimization goals
    \item Some overlap in claim scope but with substantially different limitations
\end{itemize}

\textbf{Annotator Rationale}: "These patents show moderate similarity in their approach to image processing, sharing common preprocessing techniques and some architectural elements. However, they diverge significantly in their core algorithms, with Patent C using a convolutional approach while Patent D employs a transformer-based method. Their application domains are distinct (medical diagnosis vs. autonomous navigation), leading to different optimization goals and performance metrics. The claim scope shows some overlap in general image processing methods but contains substantially different limitations reflecting their distinct applications."

\subsubsection{Example 3: Low Similarity (Score 1)}
\textbf{Patent Pair}: PAtent E and Patent F (blockchain system vs. wireless communication protocol)

\textbf{Key Differences}:
\begin{itemize}
    \item Completely different technical domains and problem spaces
    \item No overlap in core technologies or methodologies
    \item Different implementation architectures and components
    \item Distinct application domains and user groups
    \item No similarity in claim scope or legal protection
\end{itemize}

\textbf{Annotator Rationale}: "These patents address fundamentally different technical domains with no meaningful overlap. Patent E describes a blockchain consensus mechanism for financial transactions, while Patent F details a wireless communication protocol for IoT devices. They employ different technologies, serve different purposes, target different users, and have no overlap in their implementation approaches. The claims seek protection for entirely unrelated inventions with no common elements that would create potential for infringement or prior art concerns."

\subsubsection{Reference Materials Provided to Annotators}
\begin{itemize}
    \item USPTO Manual of Patent Examining Procedure (MPEP) Section 2141
    \item European Patent Office Guidelines G-VII, 5.1
    \item Prior art search methodology guides
    \item IPC classification reference materials
\end{itemize}

\subsection{Quality Control Process}
To ensure annotation quality:
\begin{itemize}
    \item Cases with score differences >2 points were assigned to additional reviewers
    \item Weekly calibration meetings were held to discuss discrepancies
    \item Patent pairs with final standard deviation >2.0 were excluded from the dataset
    \item Senior patent experts provided final validation of all annotations
\end{itemize}

\section{Appendix G: Prompts for Baseline Methods}
\label{app:baseline_prompts}

To ensure fair comparison and reproducibility, we provide the exact prompts used for the baseline methods, including Chain-of-Thought (CoT) and Few-shot prompting.

\subsection{Prompt for Chain-of-Thought (CoT)}
The CoT prompt encourages the model to generate intermediate reasoning steps before predicting the final score.

\noindent Prompt: \textit{You are an expert patent analyst. Your task is to evaluate the similarity between two patent documents, Patent A and Patent B.
\newline
Input:
\newline
- Patent A: [Title, Abstract, Claims]
\newline
- Patent B: [Title, Abstract, Claims]
\newline
Instructions:
\newline
1. Analyze the technical field, core innovation, and claim scope of both patents.
2. Compare the similarities and differences step-by-step.
3. Based on the reasoning, determine a similarity score between 0.0 (completely different) and 1.0 (identical).
\newline
Output the result in the following format:
\newline
Reasoning: [Your step-by-step analysis]
\newline
Score: [numerical score]}

\subsection{Prompt for Few-shot Prompting}
For Few-shot prompting, we utilized a 3-shot setting. We selected three representative patent pairs with expert-annotated scores (Low, Medium, High) to guide the model.

\noindent Prompt: \textit{Evaluate the similarity between two patents on a scale of 0.0 to 1.0. Here are three examples:
\newline
Example 1:
\newline
Patent A: [Content of Patent A1]
\newline
Patent B: [Content of Patent B1]
\newline
Score: 0.2 (Low Similarity)
\newline
\newline
Example 2:
\newline
Patent A: [Content of Patent A2]
\newline
Patent B: [Content of Patent B2]
\newline
Score: 0.5 (Medium Similarity)
\newline
\newline
Example 3:
\newline
Patent A: [Content of Patent A3]
\newline
Patent B: [Content of Patent B3]
\newline
Score: 0.9 (High Similarity)
\newline
\newline
Task:
\newline
Patent A: [Target Patent A Content]
\newline
Patent B: [Target Patent B Content]
\newline
Score:}

\clearpage
\section{Appendix H : Detailed High-Error Case Analysis}  
\label{appendix:high_error_cases}  

\renewcommand{\arraystretch}{1.3} 

\begin{table}[h]
\centering
\small
\caption{Representative Cases of Model Mispredictions}
\label{tab:high_error_cases}
\begin{tabularx}{\textwidth}{lXccc}
\toprule
\textbf{Error Type} & \textbf{Patent Pair} & \textbf{Ground Truth} & \textbf{Model Prediction} & \textbf{Error} \\
\midrule
\multirow{2}{*}{Underestimation} & \textit{METHOD FOR PLATINUM RECOVERY} (C22B300) & 0.833 & 0.53 & 0.303 \\
& \textit{METHOD FOR PLATINUM RECOVERY} (C22B300) & & & \\
\midrule
\multirow{3}{*}{Overestimation} & \textit{METHOD FOR APPLYING INK MARKINGS} (G06K1502) & 0.250 & 0.50 & 0.250 \\
& \textit{METHOD FOR DETERMINING QUALITY OF MARKINGS} (G06T700) & & & \\
\midrule
\multirow{2}{*}{Functional Differences} & \textit{DISC BRAKE} (F16D6500) & 0.750 & 0.48 & 0.270 \\
& \textit{DISC BRAKE} (F16D65097) & & & \\
\midrule
\multirow{2}{*}{Terminology} & \textit{DNA POLYMERASES WITH INCREASED 3'-MISMATCH DISCRIMINATION} (C12N912) & 1.000 & 0.70 & 0.300 \\
& \textit{DNA POLYMERASES WITH INCREASED 3'-MISMATCH DISCRIMINATION} (C12N912) & & & \\
\midrule
\multirow{2}{*}{Overestimation} & \textit{Audio Content Auditioning by Playback Device} (G06F316) & 0.500 & 0.24 & 0.260 \\
& \textit{Systems and Methods for Automatically Generating Audio Content} (H04L2906) & & & \\
\bottomrule
\end{tabularx}
\end{table}

\end{document}